\documentclass{article}
\usepackage{ijcai19}

\usepackage{times}
\usepackage{soul}
\usepackage{url}
\usepackage[hidelinks]{hyperref}
\usepackage[utf8]{inputenc}
\usepackage[small]{caption}
\usepackage{graphicx}
\usepackage{amsmath}
\usepackage{booktabs}
\usepackage{algorithm}
\usepackage{algorithmic}
\title{Knowledge-Based Learning through Feature Generation}

\author{
    Michal Badian \and Shaul Markovitch
   \affiliations
    Department of Computer Science, \\
   Technion - Israel Institue of Technology, \\
    32000 Haifa, Israel \emails
   michalbad6@gmail.com, shaulm@cs.technion.ac.il
}

\begin{document}

\maketitle

\begin{abstract}
Machine learning algorithms have difficulties to generalize over a small set of examples.  Humans can perform such a task by exploiting vast amount of background knowledge they possess. One method for enhancing learning algorithms with external knowledge is through feature generation.  In this paper, we introduce a new algorithm for generating features based on a collection of auxiliary datasets.  We assume that, in addition to the training set, we have access to  additional datasets.  Unlike the transfer learning setup, we do not assume that the auxiliary datasets represent learning tasks that are similar to our original one.  The algorithm finds features that are common to the training set and the auxiliary datasets.  Based on these features and examples from the auxiliary datasets, it induces predictors for new features from the auxiliary datasets.  The induced predictors are then added to the original training set as generated features.  Our method was tested on a variety of learning tasks, including text classification and medical prediction, and showed a significant improvement over using just the given features.
\end{abstract}

\section{Introduction}

Machine learning algorithms  attempt to learn a predictor by generalizing over 
a given set of tagged examples. In recent years, we have seen a significant progress
in solving complex learning problems using algorithms based on Artificial Neural Networks (ANN). 
These algorithms  perform well when a large set of
tagged examples is available, but their performance rapidly declines 
when the size of the training set decreases.

While machines find the task of inducing over a small set of examples
problematic, humans are usually very good at such tasks.  A human
is capable, in many cases, to induce over an extremely small set of examples
using extensive background knowledge.
With the increasing availability of large-scale knowledge bases, 
machine learning would have significantly benefited if it could have 
utilized a similar knowledge-based approach.

There are three major approaches for using extra knowledge to
overcome the lack of examples.  In the 80's, several research groups had
persuaded an approach called \emph{Explanation Based Learning} 
\cite{Mitchell:1986:EBG,DeJong:1986:ALO},
where even a single example can be used for induction.  
The method works by generating and generalizing
a proof that explains the example 
using background knowledge in the form of a set of logical assertions.
This approach is still not practical in most cases due to the lack of such logic-based knowledge bases.

A second approach for solving a learning task with a small training set
is  \emph{transfer learning} \cite{Transfer_Survey}.
This method assumes that, in addition to the original training set, there
is another, usually larger, a set of examples of a related learning task.
The above two approaches are limited in the type of background knowledge they can exploit for enhancing the learning process.

A third approach uses external knowledge to generate new features.   
Two prominent works in this direction are Explicit Semantic Analysis \cite{Gabrilovich_shaul}, where Wikipedia-based conceptual features are generated and added to 
textual examples, and Word2Vec \cite{word2vec} where  latent concepts, based on a large corpus, are generated and used as features for text classification.  These two methods proved to be very effective but they
are applicable for text-based learning tasks only.

With the significant growth in the usage of machine learning for various
problems domains, the amount of knowledge encoded in datasets has significantly increased.  In addition, we are seeing a considerable proliferation in the
volume of user-contributed knowledge encoded in relational form. 
These bodies of external knowledge could have made a great source 
 for enhancing learning algorithms. 

In this work we present a general algorithm that enhances machine learning algorithms
with features that are automatically generated using such external knowledge sources.
Given a learning task with a specific training set, our approach takes a set of external datasets and looks for matching features.
For each external dataset, these are its set of features that are in common 
with the original features of the training set.  We use these features to learn
a classifier or a regressor that predicts the values of other features in the external dataset.
The learned predictors
are then applied to the original dataset to get newly generated features.

To understand this process, let us look at the following
illustrative example.  Assume we are given
a small training set for predicting the risk of a patient to have Alzheimer disease.
The set of features for this dataset includes, among others,  \emph{BMI}, \emph{blood pressure} and \emph{HDL cholesterol}.  We  also have  access to a much larger dataset for
evaluating the risk of a patient to have diabetes 2 that also includes these three features.  Our method
builds a new secondary learning problem, where the examples
are those of the diabetes dataset, the features are the three common features, and the target concept is the risk of having diabetes.
The resulting classifier  predicts the risk of diabetes based on these three features.
This  classifier will then be applied to the original dataset to
yield values for a newly generated feature -- the risk of having diabetes.
The generated feature is likely to enhance the ability of the learning algorithm to 
induce a predictor for the original task, predicting the risk of Alzheimer, as diabetes is one of the risk factor for this disease.c

It is important to note that our method does not compete with alternative feature generation
techniques but complements them.  It is perfectly feasible to have two feature generation processes 
work in tandem, where one generates features using auxiliary datasets with our method, while 
the other generates additional features, for example, using a text corpus with Word2Vec.  Our method can also be integrated with combinational feature generation methods -- we first apply our algorithm to generate new features based on external knowledge, and then apply a combinational method on
the enhanced pool of features.

\section{Problem definition}
We assume that we are given a training set to learn from, and an additional set of auxiliary datasets, either labeled or unlabeled.
More formally, we define a \emph{dataset} $D$ as a pair $\left \langle O,F \right \rangle$,
where $O$ is a set of objects
and $F$ is a set of features such that
$\forall f \in F : O \subseteq Domain(f)$.  We sometimes denote the set of features 
of a dataset $D$  by $F(D)$.

Given a dataset $D = \left \langle O,F \right \rangle$, 
a \emph{labeled dataset} with respect to $D$ and
 a target function $f^*$, is defined as $ \left \langle E,F \right \rangle$,
where $E=\{ \left \langle o,f^*(o) \right \rangle | o \in O\}$.
We sometimes denote the set of objects of a dataset $D$ by $O(D)$.

A learning algorithm takes a labeled dataset ${D_t= \left \langle E_t,F_t \right \rangle}$ 
and yields a classifier $f_c : O \rightarrow Range(f^*)$.

A Feature Generation Task (FGT) is a pair $(D_t,{\cal A})$, where
$D_t=\left \langle E_t,F_t \right \rangle$ is a labeled training set, and 
${\cal A} = \{D_1,...,D_k\}$, where $D_i$ is either labeled or unlabeled, is a set of auxiliary datasets.

A knowledge-based feature generation algorithm takes a FGT as an input and generates a set of features $F_g$
over $O(D_t)$. We call the dataset ${D_t^g= \left \langle E_t,F_t \cup F_g\right \rangle}$ the 
\emph{enhanced dataset}. We name the main learning algorithm, 
that will be applied to $D_t^g$, the \emph{primary} learning algorithm, and denote it by $L_p$.

Our goal is to exploit the auxiliary datasets to improve 
the quality of the induced predictor.  We propose to do so via feature generation 
as described in the next section.

\section{The algorithm}
\label{subsection:algorithm}

In this section we present a feature generation framework that can
exploit additional datasets and inject the knowledge embedded in these
datasets into the learning process.  
We start with the basic algorithm and continue with extensions.

\subsection{The Basic Algorithm}
For each auxiliary dataset $D_a$, our algorithm first finds the set of common features with the training set $F^{\cap} = F(D_a) \cap F(D_t)$, and the set of features 
$F^{-} = F(D_a) - F(D_t)$ that appears only in the auxiliary dataset.

For each feature $f \in F^{-}$, the algorithm  constructs a new learning problem,
called a \emph{secondary learning task}, where the objects are $O(D_a)$, the target function
is $f$, and the features are $F^{\cap}$. A \emph{secondary} learning algorithm, denoted by $L_s$, 
is then applied to the secondary learning task.
Note that $L_p$ and $L_s$ are not necessarily related.

The learning process results
in a predictor $f_g$.  This is a predictor (a classifier or a regressor) that is based only on the
features in $F^{\cap}$ and therefore can be added to $F(D_t)$ as a newly generated feature.
Before we add it to the growing set of generated features, we first apply a wrapper selector\cite{Kohavi:1997:WFS} and add it only if it has positive utility.  The pseudocode for the algorithm is listed below. $CV$ stands for cross validation estimation.

\begin{algorithm}[H]
\caption{Knowledge-Based Feature Generation Algorithm}
\label{alg:algorithm}
\textbf{KBFG($D_t$, $D_a$)}\\
\textbf{Input}: $ \left \langle E_t,F_t \right \rangle$, $ \left \langle O_a,F_a \right \rangle$
\begin{algorithmic}[1] 
\STATE $F_g \leftarrow \{\}$
\STATE $F^{\cap} \leftarrow F_t \cap F_a$
\FOR{$f \in F_a - F^\cap$}
\STATE $\hat{E} \leftarrow \{ \left \langle o,f(o) \right \rangle | o \in O_A \} $
\STATE $f_g \leftarrow L_s(\hat{E}, F^\cap)$
\IF {$CV(F_t \cup F_g \cup \{f_g\}) > CV(F_t \cup F_g)$}
\STATE $F_g \leftarrow F_g \cup \{f_g\}$
\ENDIF
\ENDFOR
\STATE \textbf{return} $F_g$
\end{algorithmic}
\end{algorithm}

\subsection{The Secondary Learning Task}
\label{subsection:secondary}
The algorithm described in the previous subsection generates new features by
creating a new learning problem and solving it.  The components of this stage are as follows:
\begin{enumerate}
    \item The objects of the new problem are the objects of the auxiliary dataset. 
    \item The target function is a feature in the auxiliary dataset that does not appear in the original dataset.  Note that the auxiliary dataset can be either labeled or unlabeled.
If it is labeled, we treat its target class as a part of the set of features.  Thus, the generated
features may include the class.  
\item In some datasets, for instance in textual ones, the set of such features is very large.  Since each feature requires a full learning process, we may want to prioritize this set. When there exists a distance function between potential features and the original dataset, it can be used for making prioritization.  In the experimental section, we describe an experiment where the auxiliary dataset is a very large text corpus (Wikipedia).  The distance function we used there is based on Word2Vec embedding.
We compute the similarity of the potential feature to the positive training examples, and the similarity to the negative ones.
The maximal similarity taken as its utility. 
    \item The features used for the secondary learning problem are those that are common to the original dataset and the auxiliary dataset.  When this set is very large we perform a feature selection.
    \item The learning algorithm produces a classifier when the target feature is nominal, and a regressor if it is continuous.  In the experiments reported here, we have used a random forest classifier (or regressor)  but any other algorithm can be used.
\end{enumerate}

\begin{figure}[t]
\includegraphics[width = \columnwidth]{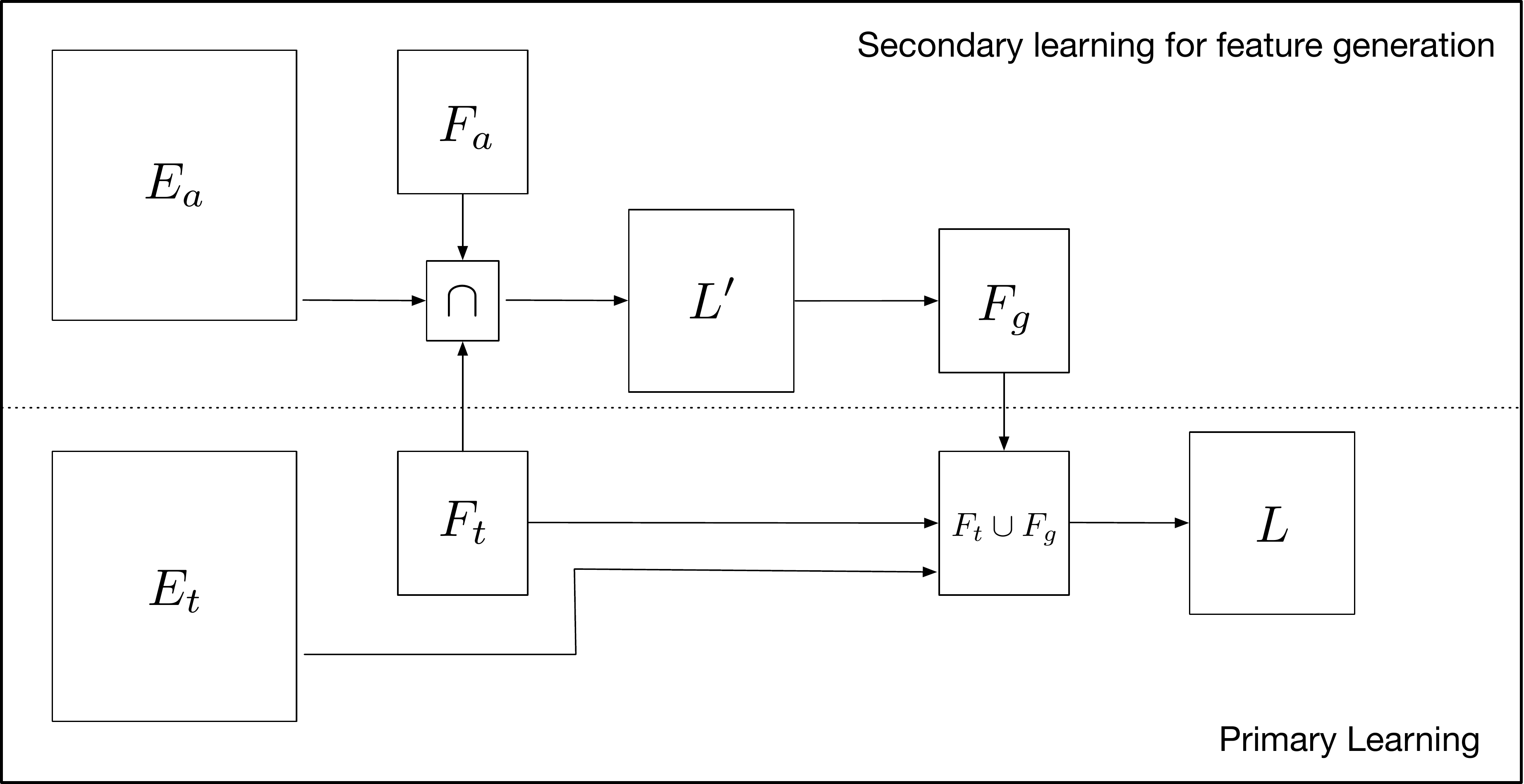}
\caption{The general architecture of the feature generation framework}
\label{fig:system}
\end{figure}

The architecture of our framework is illustrated in Figure \ref{fig:system}

\subsection{Feature Matching}
Our algorithm assumes that $F^{\cap} = F_t \cap F_a$ is computable.
One way to achieve this is to manually specify a feature matching table by the user.
Otherwise, we need an automatic way to do the matching.
If the names of the features are given explicitly, such as in the UCI collection \cite{Dua:2017},
and the same names are used in the training dataset and the auxiliary dataset (or the same keys such in SQL databases), the matching is straightforward. 

Otherwise, we define some similarity measure between features, and match them only if their similarity is above some threshold.
If the names are only slightly different, we can use stemming and Levenshtein distance
to find matching.  If the names differ, we can use embedding such as Word2Vec 
to estimate the similarity between pairs of feature names.  When  descriptions for features are given,
such as the case in many UCI datasets, we embed their description as well.

If feature names are not given, we can estimate the similarity of two categorical or numerical features by measuring the distance between the distributions of their values, using common distribution distance measures.

\subsection{Recurrent Feature Generation}
Once algorithm \ref{alg:algorithm} generates a feature $f_g$ and add it to the training set,
the intersection set $F^{\cap}$ now includes a new member. By the way we defined the process, we now have a state where $f_g$ is in the enhanced training features, and the feature it approximates, $f$, exists in the auxiliary set of features.  Therefore, we are able to match these two features 
and expand the intersection set, which may lead to an improvement in the induction of additional generated features. The recurrent version of our algorithm includes one additional line.
After the last step of the $for$ loop, we add the following statement: 
$F^{\cap} \leftarrow F^{\cap}\cup \{f_g\}$.

\subsection{Using Multiple Auxiliary Datasets}

When we get a set of auxiliary datasets, we can call our basic feature-generation algorithm, $KBFG$, for each auxiliary dataset sequentially.  One drawback of this approach, is that the wrapper feature selector evaluates each generated feature only with respect to the local set of features generated for the particular dataset, and not with respect to the global generated set of features. 
Therefore, we propose to add an extra feature selection stage that will be performed on the accumulated set of generated features after processing all the auxiliary datasets.

We considered the wrapper approach, which iteratively evaluates the contribution of each feature with respect to the current set and adds the one with the best utility.  This, however, has quadratic complexity (with respect to $n=|\bar{F_g}|$, the number of generated features). With thousands of generated features, it may require millions of cross-validation experiments.

An alternative cheap method  for selection is the filtering approach, where we compute some statistical measure for each feature, such as information gain, and select only the top features.  This method, however, evaluates each feature independently of the others.  

We have therefore designed a hybrid feature selection approach that performs an order of $n$ cross validation operations.  We first sort the features  $\bar{F_g}$ by their information gain, starting with the highest.  Then we start to add them to $F_t$ one by one, evaluating their relative utility using cross validation.
The pseudo-code for the $KBFG^*$ for feature generation with multiple-auxiliary datasets is listed in Algorithm \ref{alg:gathering_algorithm}.

There are several other considerations when using multiple datasets:
\begin{enumerate}
    \item Obviously, there is no point in using an auxiliary dataset if the intersection set is empty.
    \item  It is possible that a feature $f$ belongs to multiple datasets and therefore
    multiple approximations of it, $F=\{f_g^1,\ldots,f_g^k\}$, will be induced.  
    One option to deal with this case is choosing 
    \[
    f'= \operatorname*{arg\,max}_{f \in F} U(f,D_t) 
    \]
    where $U$ is any feature evaluation method such as wrapper or information gain.
    Alternatively, we can form a committee $\{f_g^1,\ldots,f_g^k\}$ that will serve as the new generated feature $f_g$.
    \item We can use the recurrent version of the algorithm with multiple auxiliary datasets.
    Assume that $f_g$ was generated using an auxiliary dataset $D_a$.  We can add this feature 
    to other auxiliary datasets, thus enhancing their expressiveness. This can be done only if
    the features used to learn it are included there.
\end{enumerate}

\begin{algorithm}[H]
\caption{The $KBFG^*$ For Feature Generation with Multiple Auxiliary Datasets}
\label{alg:gathering_algorithm}
\textbf{KBFG$^*$($D_t$, ${\cal A}$)}\\
\textbf{Input}: $ \left \langle E_t,F_t \right \rangle$, ${\cal A}$
\begin{algorithmic}[1] 
\STATE $\bar{F_g} \leftarrow \{\}$
\FOR{$\left \langle O_a,F_a \right \rangle \in {\cal A}$}
\STATE $F_g$ = $KBFG ( \left \langle E_t,F_t \right \rangle \, \left \langle O_a,F_a \right \rangle)$
\STATE $\bar{F_g} \leftarrow \bar{F_g} \cup F_g$
\ENDFOR
\STATE Sort $\bar{F_g}$ by information gain in descending order
\STATE $G \leftarrow \{\}$
\FOR{$f_g \in \bar{F_g}$}
\IF {$CV(L_p,F_t \cup G \cup \{f_g\}) > CV(L_p,F_t \cup G)$}
\STATE $G \leftarrow G \cup \{f_g\}$
\ENDIF
\ENDFOR
\STATE \textbf{return} $G$
\end{algorithmic}
\end{algorithm}

\begin{table*}[tb]
\centering
\begin{tabular}{|l|r|l|r|r|r|}
\hline
Training Set & \#Examples & Auxiliary dataset & \#Examples & AVG $|F^\cap|$ &  AVG $|F^-|$ \\
\hline
IMDB$^{0.25}$ & 225 & AMAZON & 1000 & 611 & 705 \\
AMAZON$^{0.25}$ & 225 & IMDB & 1000 & 611 & 705 \\
IMDB$^{0.25}$ & 225 & YELP & 1000 & 642 & 873 \\
YELP$^{0.25}$ & 225 & IMDB & 1000 & 642 & 873 \\
AMAZON$^{0.25}$ & 225 & YELP & 1000 & 494 & 822 \\
YELP$^{0.25}$ & 225 & AMAZON & 1000 & 494 & 822 \\
Pima$^{0.25}$ & 172 & Breast Cancer & 116 & 4 & 5 \\
Breast Cancer$^{0.25}$ & 26 & Pima & 768 & 4 & 4 \\
ILPD$^{0.25}$ & 130 & Hepatitis & 80 & 5 & 14 \\
Hepatitis$^{0.25}$ & 18 & ILPD & 579 & 5 & 5 \\
Cylinder$^{0.25}$ & 61 & Cylinder & 265 & 16 & 21 \\
SPECTF$^{0.25}$ & 40 & SPECTF & 137 & 18 & 24 \\
QSAR$^{0.25}$ & 117 & QSAR & 531 & 17 & 22 \\
Z-Alizadeh$^{0.25}$ & 32 & Z-Alizadeh & 159 & 24 & 32 \\
Cardio$^{0.25}$ & 235 & Cardio & 1077 & 9 & 11 \\
\hline
\end{tabular}
\caption{The FGTs in our experiments. $\alpha = 0.25$, $F_a = f^\cap \cup f^-$}
\label{table:datasets_ftg}
\end{table*}


\section{Empirical evaluation}
We have tested our framework using an extensive collection of training sets with associated auxiliary datasets.

\subsection{Experimental Methodology}
We have used 12 original datasets that served as the basis for our feature generation experiments\footnote{Each dataset was used after preprocessing and normalization.}.  The first three are textual datasets of reviews tagged with their sentiment.
The rest are datasets taken from the UCI \cite{Dua:2017} and Kaggle repositories.

\subsubsection{Creating Feature Generation Tasks}
A feature generation task (FGT) is a pair of a training set and a set of auxiliary datasets.
In all experiments except those described in \ref{multipleDatasets}, we always use one auxiliary dataset.
One major step in implementing our framework is finding matching between the features of
the training set and those of the auxiliary dataset.  For the text classification tasks,
was have just matched the stemmed words. Thus, we created 6 feature generation tasks out of the 3 sentiment analysis datasets.  \emph{Pima}  and  \emph{Breast Cancer} datasets are medical 
records containing common features such as Glucose and BMI, and therefore could be paired for feature generation.  So are the \emph{Indian Liver} and \emph{Hepatitis } datasets, with common features such as Bilirubin and Age.  

Since we needed more FGTs, we have designed a method of creating an FGT out of one dataset.
The examples of the datasets are randomly partitioned to create two disjoint sets of examples, one
used for the training and one for the auxiliary dataset.  The feature set is also carefully partitioned
so that we can control the size of the intersection.

  Let $ D = \left \langle E,F \right \rangle$ be a labeled dataset:
\begin{enumerate}
\item Randomly partition $E$ into two nearly equal sized sets $E_1$ and $E_2$.
\item Randomly select a subset from $F$ of size  $\mu_1{|F|}$ ($\mu_1=\frac{1}{3}$ in our experiments)
and mark it as the intersection set, denoted by $F^{\cap}$.
\item Randomly partition $F - F^{\cap}$ into two sets: $F_1$ of size  $\mu_2|F - F^{\cap}|$ and $F_2$ ($\mu_2=\frac{2}{3}$ in our experiments).
\item The training set will be $\left \langle E_1, F_1 \cup F^{\cap}\right \rangle$
and the auxiliary dataset $\left \langle E_2, F_2 \cup F^{\cap}\right \rangle$.
\end{enumerate}

\subsubsection{Testing Protocol}
One prominent motivation for our work is to overcome
the problem of lack of examples by using external knowledge.
Therefore we have followed a special testing protocol 
that simulates such scenarios by reducing the number of examples in the training set. 

Given an FGT $\left \langle \left \langle E_t, F_t \right \rangle,
\left \langle E_a, F_a \right \rangle \right \rangle$:
\begin{enumerate}
\item Partition $E_t$ into $k$ partitions (10 in our experiments) with testing sets $T_1, \ldots, T_k$ and corresponding training sets $\left \langle E_t - T_i, F_t \right \rangle$.
\item For each test set $T_i$ and a set of examples $B_i = E_t - T_i$:
\begin{enumerate}
\item Reduce  the number of examples in $B_i$ by randomly selecting 
a subset $B_i^\alpha$ of size $\alpha \cdot | B_i |$ ($\alpha = 0.25$ in our experiments).
\item Generate a set $G_i$ of new features based on the auxiliary dataset.
\item Sort $G_i$ by information gain with respect to $E_t$. For the textual datsets, in which the size of $F^-$ is large, we select the best $50$.
\item For each $f$ in the sorted $G_i$, test if it has a positive utility value by the wrapper method, and add it to $F_t$ if it does.  We 
denote the enhanced feature set by ${\hat F_t}$ .
\item Learn from $\left \langle B_i^\alpha, {\hat F_t}  \right \rangle$.
\item Test on $T_i$.
\end{enumerate}
\item Repeat step 2 for each $L_p$ used in the experiments.
\end{enumerate}

The list of FGTs used for our experiments is shown in Table \ref{table:datasets_ftg}. The superscript next to each dataset stands for the $\alpha$ parameter used to reduce the number of examples.

\begin{table*}[tb]
\centering
\begin{tabular}{|l|l|r|r|r|r|r|r|}
\hline
Training Set & Auxiliary dataset & DT &  DT + FG  & Diff DT & SVM &  SVM + FG  & Diff SVM \\
\hline
IMDB$^{0.25}$ & AMAZON & 0.566 & 0.621 & \textbf{+0.055} & 0.695 & 0.709 & +0.014 \\
AMAZON$^{0.25}$ & IMDB & 0.542 & 0.615 & \textbf{+0.073} & 0.732 & 0.735 & +0.003 \\
IMDB$^{0.25}$ & YELP & 0.539 & 0.610 & \textbf{+0.071} & 0.691 & 0.711 & \textbf{+0.020} \\
YELP$^{0.25}$ & IMDB & 0.554 & 0.612 & \textbf{+0.058} & 0.695 & 0.710 & \textbf{+0.015} \\
AMAZON$^{0.25}$ & YELP & 0.593 & 0.672 & \textbf{+0.079} & 0.720 & 0.749 & \textbf{+0.029} \\
YELP$^{0.25}$ & AMAZON & 0.553 & 0.666 & \textbf{+0.113} & 0.696 & 0.717 & \textbf{+0.021} \\
Pima$^{0.25}$ & Breast Cancer & 0.751 & 0.828 & \textbf{+0.077} & 0.576 & 0.703 & \textbf{+0.127} \\
Breast Cancer$^{0.25}$ & Pima & 0.721 & 0.738 & +0.017 & 0.602 & 0.688 &  \textbf{+0.086} \\
ILPD$^{0.25}$ & Hepatitis & 0.660 & 0.694 & \textbf{+0.034} & 0.550 & 0.706 & \textbf{+0.156} \\
Hepatitis$^{0.25}$ & ILPD & 0.700 & 0.774 & +0.074 & 0.790 & 0.840 & +0.050 \\
Cylinder$^{0.25}$ & Cylinder & 0.633 & 0.692 & +0.059 & 0.578 & 0.610 & +0.032 \\
SPECTF$^{0.25}$ & SPECTF & 0.615 & 0.721 & \textbf{+0.106} & 0.677 & 0.812 & \textbf{+0.135} \\
QSAR$^{0.25}$ & QSAR & 0.686 & 0.732 & +0.046 & 0.726 & 0.791 & \textbf{+0.065} \\
Z-Alizadeh$^{0.25}$ & Z-Alizadeh & 0.621 & 0.718 & \textbf{+0.097} & 0.619 & 0.706 & \textbf{+0.087} \\
Cardio$^{0.25}$ & Cardio & 0.790 & 0.816 & \textbf{+0.026} & 0.757 & 0.804 & \textbf{+0.047} \\
\hline
\end{tabular}
\caption{The effect of feature generation on the performance of two learning algorithms: Decision Tree With 
pruning and linear SVM.}
\label{table:datasets_results}
\end{table*}

\begin{table*}[tb]
\centering
\begin{tabular}{|l|l|r|r|r|r|r|r|}
\hline
Training Set & Auxiliary dataset & 3NN &  3NN + FG  & Diff 3NN & MLP &  MLP + FG  & Diff MLP \\
\hline
IMDB$^{0.25}$ & AMAZON & 0.573 & 0.610 & +0.037 & 0.709 & 0.741 &  \textbf{+0.032} \\
AMAZON$^{0.25}$ & IMDB & 0.609 & 0.655 & \textbf{+0.046} & 0.752 & 0.771 &  \textbf{+0.019} \\
IMDB$^{0.25}$ & YELP & 0.552 & 0.587 & \textbf{+0.035} & 0.715 & 0.738 &  \textbf{+0.023} \\
YELP$^{0.25}$ & IMDB & 0.567 & 0.607 & \textbf{+0.040} & 0.733 & 0.747 &  \textbf{+0.014} \\
AMAZON$^{0.25}$ & YELP & 0.609 & 0.659 & \textbf{+0.050} & 0.739 & 0.773 &  \textbf{+0.034} \\
YELP$^{0.25}$ & AMAZON & 0.593 & 0.659 & \textbf{+0.066} & 0.724 & 0.749 &  \textbf{+0.025} \\
Pima$^{0.25}$ & Breast Cancer & 0.673 & 0.729 & \textbf{+0.056} & 0.568 & 0.664 &  \textbf{+0.096} \\
Breast Cancer$^{0.25}$ & Pima & 0.531 & 0.548 & +0.017 & 0.547 & 0.608 & +0.061 \\
ILPD$^{0.25}$ & Hepatitis & 0.646 & 0.675 & +0.029 & 0.658 & 0.727 & \textbf{+0.069} \\
Hepatitis$^{0.25}$ & ILPD & 0.814 & 0.840 & +0.026 & 0.795 & 0.837 & +0.042 \\
Cylinder$^{0.25}$ & Cylinder & 0.547 & 0.570 & +0.023 & 0.563 & 0.582 & +0.019 \\
SPECTF$^{0.25}$ & SPECTF & 0.635 & 0.729 & \textbf{+0.094} & 0.576 & 0.710 &  \textbf{+0.134} \\
QSAR$^{0.25}$ & QSAR & 0.706 & 0.762 & \textbf{+0.056} & 0.714 & 0.703 & -0.011 \\
Z-Alizadeh$^{0.25}$ & Z-Alizadeh & 0.575 & 0.711 & \textbf{+0.136} & 0.590 & 0.728 & \textbf{+0.138} \\
Cardio$^{0.25}$ & Cardio & 0.708 & 0.799 & \textbf{+0.091} & 0.710 & 0.773 & +0.063 \\
\hline
\end{tabular}
\caption{The effect of feature generation on the performance of two learning algorithms: KNN with K=3 and Multi-layer perceptron}
\label{table:datasets_results_two}
\end{table*}


\subsection{ Performance With the Generated Features}
We performed a set of experiments to test the utility of our feature generation algorithm.
Each experiment involved generating a set of features with a given FGT, and testing 
the enhanced feature set with various learning algorithms, including Decision Tree, 
Linear SVM, KNN (N=3) and Multilayer perceptron.  We used the implementation in \emph{sklearn} with the default parameters.

Tables \ref{table:datasets_results} and \ref{table:datasets_results_two} show the results achieved.  For each learning algorithm
the tables present the performance on the testing set without and with the generated features.
The third column shows the added accuracy.  
We mark by bold differences that are statistically significant by paired t-test with $p < 0.05$.

We can see that the generated features significantly enhance the performance in almost all cases.
Note that the improvements are orthogonal to potential other improvements that may be achieved
by feature-combination algorithms or by feature generation algorithms that use other external resources.

\subsection{Experimenting with a Very Large Auxiliary Dataset}
We have performed another experiment with textual learning task, where the auxiliary
dataset is the whole Wikipedia corpus.  We consider each Wikipedia article as one object.
Using such a large dataset is problematic.  The Bag-O-Words matrix is, after filtering,
of size $850,000 \times 700,000$ approximately.  As the training dataset contains only a few
thousands of words (the size of $F_t$), the size of $F^- = F_a - F_t$ is also of size
of almost $850,000$.

As we cannot afford initiating $850,000$ learning processes, we applied the similarity-based
prioritization as described in subsection \ref{subsection:secondary}.  Specifically, 
we have used Word2Vec distance to select the $5,000$ features closest to the centroid of the
positive examples of the training set, and another $5,000$ features closest to the negative one.

The results are shown in Table \ref{table:datasets_results_wiki}.  We can see that the new features
added about $5\%$ in accuracy (statistically significant with $p < 0.05$) with both datasets tested.

\begin{table*}[tb]
\centering
\begin{tabular}{|l|l|r|r|r|}
\hline
Training Set & Auxiliary dataset & DT &  DT + FG  & Diff DT \\
\hline
IMDB$^{0.25}$ & Wikipedia & 0.593 & 0.640 & \textbf{+0.047} \\
AMAZON$^{0.25}$ & Wikipedia & 0.690 & 0.742 & \textbf{+0.052}\\
\hline
\end{tabular}
\caption{The results of generating word-based features using the Wikipedia corpus. The improvement (almost 
5\%) is statistically significant with $p < 0.05$.}
\label{table:datasets_results_wiki}
\end{table*}





\subsection{Experimenting with multiple datasets}
\label{multipleDatasets}
We have designed a feature generation task that involve multiple auxiliary datasets.
\begin{enumerate}
\item Training set: YELP. 
\item Auxiliary datasets: IMBD and AMAZON.
\end{enumerate}

We have applied our $KBFG^*$ Algorithm (see Algorithm \ref{alg:gathering_algorithm})  to the task described.  For comparison, we have also tested the basic $KBFG$ algorithm with each of the auxiliary datasets separately.
We ran each of the 3 scenarios until 50 generated features passed the wrapper filter.  

The experiment was stopped after each 5 features were accepted and the accuracy of the classifier induced with the added features was measured as described in the testing protocol. That is, the pool of $50$ ordered features, was split into 10 groups of 5 items per each. Each point in the graphs, is an addition of the next ranked group of $5$ features, where the zero point represent the original set of features (i.e., without any additional features).

It is important to mention that the accuracy was measured using the same folds. That is, the only different is the features added using the single or multiple auxiliary. The number of the features and training set examples, are preserved.

\begin{figure}[p]
\includegraphics[width=
\columnwidth]{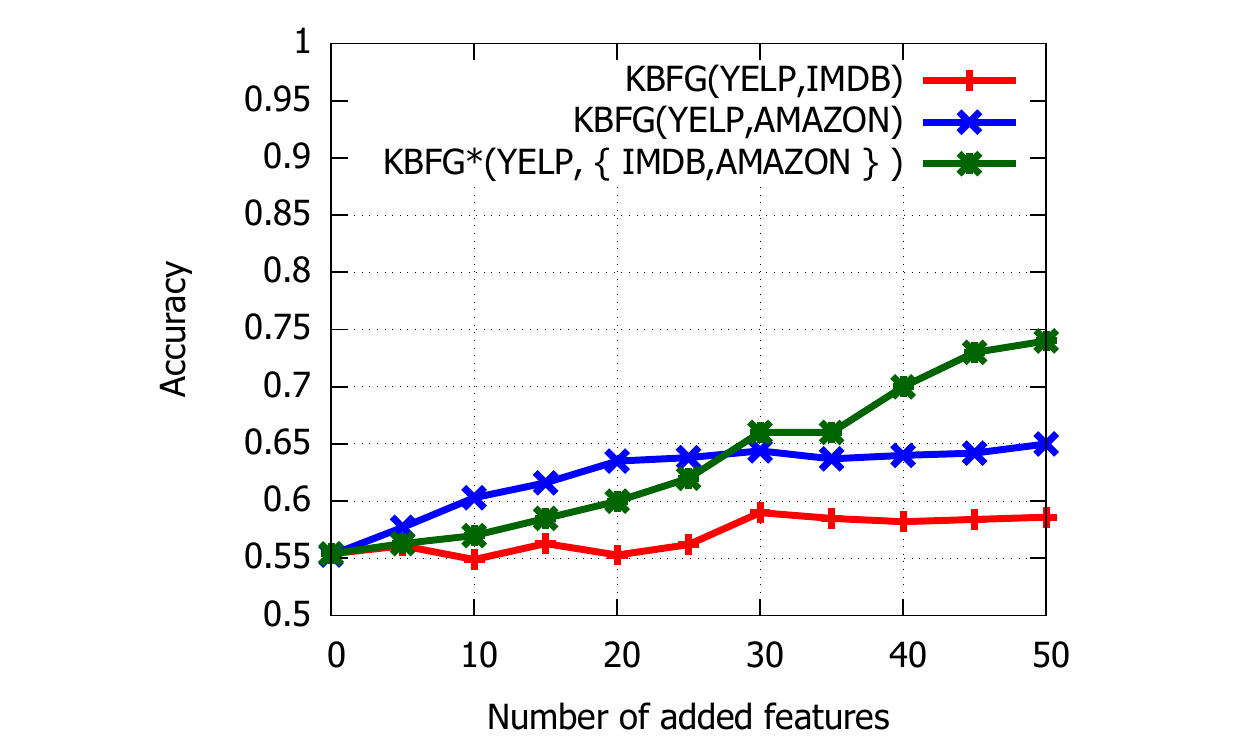}
\centering
\caption{The effect of the addition of features generated in our algorithm, on the accuracy achieved by decision tree with
pruning, on the YELP dataset.}
\label{fig:graph_multiple_yelp}
\end{figure}

The graphs in figure \ref{fig:graph_multiple_yelp} show the average accuracy of a classifier induced by decision tree with pruning with the added generated features, on the YELP dataset, for the 3 scenarios.
We can see that  the $KBFG^*$ algorithm had significant advantage over  $KBFG$ with each of the  single auxiliary dataset (about 10\% difference in accuracy).

\section{Related work}

Many feature generation methods have been developed in an attempt
to construct new features that better represent the target concept.
 The most common approach for feature generation (feature engineering) is by constructing new features that
are combinations of the given ones.
The early GALA algorithm \cite{gala} utilizes a set of logical operators to combine Boolean attributes. Another example is the LFC \cite{ragavan1993complex} algorithm that combines binary features through the use of logical operators such as $\vee , \neg$.

The CITRE algorithm \cite{citre} constructs new features by combining basic features using decision trees.
The FICUS algorithm \cite{Markovitch:2002:FGU} presents a general framework that, given a set of constructing functions, applies them over existing features, included generated ones. The more recent ExploreKit algorithm \cite{exploreKit} uses a similar technique.  The FEADIS \cite{feadis} algorithm and the Deep Feature Synthesis algorithm \cite{7344858} form new features using mathematical functions.  
The work of \cite{Tran2016} uses genetic programming (GP) with an
embedded approach to construct and select new features for binary classification problems.

Cognito \cite{Cognito} performs automatic feature engineering by exploring various feature construction choices in a hierarchical and non-exhaustive approach using a greedy strategy. \citeauthor{DBLP:journals/corr/abs-1709-07150} [\citeyear{DBLP:journals/corr/abs-1709-07150}] used reinforcement learning to explore the feature transformation graph.
The AutoLearn algorithm \cite{autolearn} identifies associations between feature pairs and uses a regularized regression model to construct a new feature from each pair. The LFE algorithm \cite{lfearticle} learns patterns between feature characteristics, class distributions, and useful transformations, in order to construct a more effective generated feature. 

All these methods are based on the assumption that merely combining existing features  
is sufficient to allow a learner that uses the combined features to yield better classifiers than by using only the original ones. 

Another group of  feature-generation methods use external knowledge to generate new features. Our algorithm belongs to this class of approaches.
Explicit Semantic Analysis (ESA) \cite{Gabrilovich_shaul} uses semantic concepts extracted from knowledge sources such as Wikipedia as features for text classification tasks. Word2Vec \cite{word2vec} generates latent concepts based on a large corpus that can also be used for text classification.

Other Algorithms for specific problem domains use domain background knowledge to construct special features. 
The bootstrapping algorithm was applied in the domain of molecular biology \cite{Bootstrapping} to generate new features by using an initial set of feature sequences produced by human experts and by using a special set of operators. The features are represented as nucleotides sequences whose structure is modified 
by biology-based operators determined by existing background knowledge.

Propositionalization approaches \cite{kramer2000bottom} rely on relational data to serve as external knowledge. They take advantage of several operators to create first-order logic predicates connecting existing data and relational
knowledge. \citeauthor{cheng2011automatedfull} [\citeyear{cheng2011automatedfull}] devised a generic propositionalization framework using linked data via relation-based queries.
FeGeLOD \cite{paulheim2012unsupervisedfull} also utilizes
linked data to automatically enrich existing datasets. FeGeLOD
uses feature values as entities and adds related knowledge to
the example, thus creating additional features.
OneBM \cite{DBLP:journals/corr/LamTSCMA17} works with multiple raw tables in a database. It joins the tables and applies corresponding predefined transformation functions on the given features types.

Some learning algorithms employ an internal process that can be viewed as feature generation. One example is the family of multilayer Neural Network (NN) algorithms which includes Deep learning algorithms \cite{lecun1998gradient}, where the activation functions of nodes in the hidden layers together with the weights of their inputs can be considered as features that had been built during the training process. Other ensemble models,  such as Random Forest (RF), use randomization to create data subsets where the new generated features are the output of decision trees applies on these data subsets.


\section{Conclusion}

One of the advantages of human learners over machine learning algorithms is their ability to exploit excessive background knowledge during the induction process.
This ability allows them to generalize even when a small set of labeled examples is given.
In this paper, we presented a novel algorithm that allows machine learning procedures to similarly exploit external knowledge via feature generation.
 
The core of our method is the realization that many datasets deal with similar types of objects,
such as people and texts.  Thus, many datasets are using common features.  Given a dataset 
to learn from, our method utilizes its  common
features with other available datasets to learn to predict additional features from these external datasets.  The predictors are then added as generated features to the original dataset.
We have shown that our method  significantly enhances the performance of existing learning algorithms.

One potential difficulty for our method is an auxiliary dataset where all features are orthogonal. In such a case the secondary learning process may not be able to learn good predictors. Note, however, that in most real datasets, the features are not fully orthogonal. Even the Naive Bayes assumption, that the features are independent given the class, does not hold in most realistic learning tasks. In addition, for labeled auxiliary datasets, we always have at least one non-orthogonal feature: the class itself.
In any case, low-quality predictors will be filtered out by our wrapper-based feature selection. Our experiments show that in all the datasets used in our experiments, the generated features improved the performance on the held-out test set.
 
The features generated using our method are not replacing the original features but added to them.
Therefore, other methods that use external knowledge can still be applied, potentially even better
as they now have access to an enhanced feature set.  Similarly, feature-combination algorithms,
such as Cognito \cite{Cognito}, LFE \cite{lfearticle} and ExploreKit \cite{exploreKit}, can now use the generated features, in addition to the original set of features, in the combination functions.

We are currently in the process of utilizing  MIMIC \cite{johnson2016mimic}, a very large medical dataset, as our auxiliary 
source for generating features for medical learning tasks.  The main challenge we are tackling is developing a good feature matching strategy.  We believe that using our method with this extensive knowledge base may significantly enhance the performance of the 
learned classifiers in this important domain.

\clearpage
\bibliographystyle{named}
\bibliography{ijcai2019}

\begin{thebibliography}{}

\bibitem[\protect\citeauthoryear{Cheng \bgroup \em et al.\egroup
  }{2011}]{cheng2011automatedfull}
Weiwei Cheng, Gjergji Kasneci, Thore Graepel, David~H. Stern, and Ralf
  Herbrich.
\newblock Automated feature generation from structured knowledge.
\newblock In {\em Proceedings of the 20th ACM international conference on
  Information and knowledge management}, pages 1395--1404. ACM, 2011.

\bibitem[\protect\citeauthoryear{Dejong and Mooney}{1986}]{DeJong:1986:ALO}
G.~Dejong and R.~Mooney.
\newblock Explanation-based learning: An alternative view.
\newblock {\em Machine Learning}, 1:145--176, 1986.

\bibitem[\protect\citeauthoryear{Dor and Reich}{2012}]{feadis}
Ofer Dor and Yoram Reich.
\newblock Strengthening learning algorithms by feature discovery.
\newblock {\em Inf. Sci.}, 189:176--190, April 2012.

\bibitem[\protect\citeauthoryear{Dua and Karra~Taniskidou}{2017}]{Dua:2017}
Dheeru Dua and Efi Karra~Taniskidou.
\newblock {UCI} machine learning repository, 2017.

\bibitem[\protect\citeauthoryear{Gabrilovich and
  Markovitch}{2009}]{Gabrilovich_shaul}
Evgeniy Gabrilovich and Shaul Markovitch.
\newblock Wikipedia-based semantic interpretation for natural language
  processing.
\newblock {\em Journal of Artificial Intelligence Research}, 34:443--498, 2009.

\bibitem[\protect\citeauthoryear{Hirsh and Japkowicz}{1994}]{Bootstrapping}
Haym Hirsh and Nathalie Japkowicz.
\newblock Bootstrapping training-data representations for inductive learning:
  {A} case study in molecular biology.
\newblock In {\em Proceedings of the 12th National Conference on Artificial
  Intelligence, Seattle, WA, USA, July 31 - August 4, 1994, Volume 1.}, pages
  639--644, 1994.

\bibitem[\protect\citeauthoryear{Hu and Kibler}{1996}]{gala}
Yuh-Jyh Hu and Dennis Kibler.
\newblock Generation of attributes for learning algorithms.
\newblock pages 806--811, 1996.

\bibitem[\protect\citeauthoryear{Johnson \bgroup \em et al.\egroup
  }{2016}]{johnson2016mimic}
Alistair~EW Johnson, Tom~J Pollard, Lu~Shen, H~Lehman Li-wei, Mengling Feng,
  Mohammad Ghassemi, Benjamin Moody, Peter Szolovits, Leo~Anthony Celi, and
  Roger~G Mark.
\newblock Mimic-iii, a freely accessible critical care database.
\newblock {\em Scientific data}, 3:160035, 2016.

\bibitem[\protect\citeauthoryear{{Kanter} and {Veeramachaneni}}{2015}]{7344858}
J.~M. {Kanter} and K.~{Veeramachaneni}.
\newblock Deep feature synthesis: Towards automating data science endeavors.
\newblock In {\em 2015 IEEE International Conference on Data Science and
  Advanced Analytics (DSAA)}, pages 1--10, Oct 2015.

\bibitem[\protect\citeauthoryear{Katz \bgroup \em et al.\egroup
  }{2016}]{exploreKit}
Gilad Katz, Eui Chul~Richard Shin, and Dawn Song.
\newblock Explorekit: Automatic feature generation and selection.
\newblock In Francesco Bonchi, Josep Domingo-Ferrer, Ricardo~A. Baeza-Yates,
  Zhi-Hua Zhou, and Xindong Wu, editors, {\em ICDM}, pages 979--984. IEEE,
  2016.

\bibitem[\protect\citeauthoryear{Kaul \bgroup \em et al.\egroup
  }{2017}]{autolearn}
Ambika Kaul, Saket Maheshwary, and Vikram Pudi.
\newblock Autolearn - automated feature generation and selection.
\newblock pages 217--226, 2017.

\bibitem[\protect\citeauthoryear{Khurana \bgroup \em et al.\egroup
  }{2016}]{Cognito}
U.~Khurana, D.~Turaga, H.~Samulowitz, and S.~Parthasrathy.
\newblock Cognito: Automated feature engineering for supervised learning.
\newblock pages 1304--1307, 2016.

\bibitem[\protect\citeauthoryear{Khurana \bgroup \em et al.\egroup
  }{2017}]{DBLP:journals/corr/abs-1709-07150}
Udayan Khurana, Horst Samulowitz, and Deepak~S. Turaga.
\newblock Feature engineering for predictive modeling using reinforcement
  learning.
\newblock {\em CoRR}, abs/1709.07150, 2017.

\bibitem[\protect\citeauthoryear{Kohavi and John}{1997}]{Kohavi:1997:WFS}
Ron Kohavi and George~H. John.
\newblock Wrappers for feature subset selection.
\newblock {\em Artificial Intelligence}, 97(1-2):273--324, December 1997.

\bibitem[\protect\citeauthoryear{Kramer and Frank}{2000}]{kramer2000bottom}
Stefan Kramer and Eibe Frank.
\newblock Bottom-up propositionalization.
\newblock In {\em ILP Work-in-progress reports}, 2000.

\bibitem[\protect\citeauthoryear{Lam \bgroup \em et al.\egroup
  }{2017}]{DBLP:journals/corr/LamTSCMA17}
Hoang~Thanh Lam, Johann{-}Michael Thiebaut, Mathieu Sinn, Bei Chen, Tiep Mai,
  and Oznur Alkan.
\newblock One button machine for automating feature engineering in relational
  databases.
\newblock {\em CoRR}, abs/1706.00327, 2017.

\bibitem[\protect\citeauthoryear{LeCun \bgroup \em et al.\egroup
  }{1998}]{lecun1998gradient}
Yann LeCun, L{\'e}on Bottou, Yoshua Bengio, and Patrick Haffner.
\newblock Gradient-based learning applied to document recognition.
\newblock {\em Proceedings of the IEEE}, 86(11):2278--2324, 1998.

\bibitem[\protect\citeauthoryear{Markovitch and
  Rosenstein}{2002}]{Markovitch:2002:FGU}
Shaul Markovitch and Danny Rosenstein.
\newblock {Feature Generation Using General Constructor Functions}.
\newblock {\em Machine Learning}, 49:59--98, 2002.

\bibitem[\protect\citeauthoryear{Matheus and Rendell}{1989}]{citre}
Christopher Matheus and Larry~A. Rendell.
\newblock Constructive induction on decision trees.
\newblock pages 645--650, 1989.

\bibitem[\protect\citeauthoryear{Mikolov \bgroup \em et al.\egroup
  }{2013}]{word2vec}
Tomas Mikolov, Ilya Sutskever, Kai Chen, Greg~S Corrado, and Jeff Dean.
\newblock Distributed representations of words and phrases and their
  compositionality.
\newblock pages 3111--3119, 2013.

\bibitem[\protect\citeauthoryear{Mitchell \bgroup \em et al.\egroup
  }{1986}]{Mitchell:1986:EBG}
T.~M. Mitchell, R.~M. Keller, and S.~T. Kedar-Cabelli.
\newblock Explanation-based generalization: {A} unifying view.
\newblock {\em Machine Learning}, 1:47--80, 1986.

\bibitem[\protect\citeauthoryear{Nargesian \bgroup \em et al.\egroup
  }{2017}]{lfearticle}
Fatemeh Nargesian, Horst Samulowitz, Udayan Khurana, Elias~B. Khalil, and
  Deepak Turaga.
\newblock Learning feature engineering for classification.
\newblock In {\em Proceedings of the 26th International Joint Conference on
  Artificial Intelligence}, pages 2529--2535. AAAI Press, 2017.

\bibitem[\protect\citeauthoryear{{Pan} and {Yang}}{2010}]{Transfer_Survey}
S.~J. {Pan} and Q.~{Yang}.
\newblock A survey on transfer learning.
\newblock {\em IEEE Transactions on Knowledge and Data Engineering},
  22(10):1345--1359, Oct 2010.

\bibitem[\protect\citeauthoryear{Paulheim and
  F{\"u}mkranz}{2012}]{paulheim2012unsupervisedfull}
Heiko Paulheim and Johannes F{\"u}mkranz.
\newblock Unsupervised generation of data mining features from linked open
  data.
\newblock page~31, 2012.

\bibitem[\protect\citeauthoryear{Ragavan \bgroup \em et al.\egroup
  }{1993}]{ragavan1993complex}
Harish Ragavan, Larry Rendell, Michael Shaw, and Antoinette Tessmer.
\newblock Complex concept acquisition through directed search and feature
  caching.
\newblock In {\em Proceedings of the Thirteenth International Joint Conference
  on Artificial Intelligence (IJCAI-1993)}, pages 946--951, 1993.

\bibitem[\protect\citeauthoryear{Tran \bgroup \em et al.\egroup
  }{2016}]{Tran2016}
Binh Tran, Bing Xue, and Mengjie Zhang.
\newblock Genetic programming for feature construction and selection in
  classification on high-dimensional data.
\newblock {\em Memetic Computing}, 8(1):3--15, Mar 2016.

\end{thebibliography}

\end{document}